# Stabilizing knowledge through standards

*A perspective for the humanities*

Laurent Romary, INRIA & HUB-IDSL

Dariah research infrastructure

## Introduction

This paper is all about a paradox. How indeed can we associate the notion of standardisation, which is all about fixing rules for a given field of knowledge, with scholarly work, which, on the contrary, is basically about departing from existing knowledge and the discovery of new concepts? It may be even more paradoxical to address the issue of standardisation in the humanities, which does not seem to rely on specific technological environments to fulfil its research endeavours.

Still, this rhetorical association is not so much a paradox if we consider that science[1] is all about sharing information between researchers, thus requiring that minimal joint practices are actually agreed upon among them. Indeed, bibliographical references for instance, have been since ages the subject of standardization activities, so that one can easily retrieve the exact source of a given citation[2]. Further, the increasing role of data in science has already encouraged some scientific communities, such as astronomers, to define coherent protocols and formats for sharing their information.

All in all, the humanities are not that far behind the other scholarly fields, but efforts remain to be made to widen the acceptance of pioneer works in digital humanities or language technology. In this context, we attempt to provide a picture of what an involvement of scholars in standardisation could look like, by eliciting both the possible conceptual and technical backgrounds. Our main objective is to contribute to extend the awareness of the scholarly community on the role of standardised data and possibly encourage more scholars to consider participating to the definition of such standards as a natural component of their mission.

## Dealing with data in science

The quick and intense evolution of information technologies has had, just as in all aspects of the human society, a very strong impact on the scholarly world. One can now observe that a new generation of digital scholars has emerged, who are now carrying out most of their research activities online, and fears are already expressed that it may become difficult to deal with the forthcoming deluge of data[3].

---

[1] We are using here the word science in a broad sense, i.e. encompassing natural, as well as social and human sciences.
[2] See for instance the guidelines of the Modern Language Association (http://www.mla.org)
[3] See "Riding the Wave: How Europe can gain from the raising tide of scientific data", report Report of the High-Level Group on Scientific Data, John Wood (Ed.)

We knew already that science is essentially a matter of information. For more than two centuries, the wide dissemination of scientific journals has allowed research results to be quickly and widely known by the corresponding scientific community. In domains like mathematics, this has become the major record of science, whereas other scientific branches have identified the importance of research data (protocols as well as results proper) as an essential element of accumulating knowledge. In the human sciences, primary sources have always been the basis for the grounding and comparison of scholarly statements.

## Lexica as a representative example in the humanities

Still, humanities cannot be taken exactly in the same manner as other scholarly fields. Indeed, the main source of information, the *primary* sources, are part of a continuum of textual documents that makes observations and conclusions be pretty much of the same nature. A simplified view on humanities research could indeed make it boil down to producing commentaries on existing textual sources, where these commentaries in turn become sources for further scholarly work.

However, the situation is by far more complex than this simplified picture and humanities scholars have over the years founded their research upon a variety of information sources, most of them becoming digital over the years. Archaeologists, as the first example that would come to mind, have always been forced to maintain huge collections of place and object descriptions to be able to identify precisely correlations between discoveries. Historians of art have had similar needs of recording the characteristics of artistic objects or building. Even more, they have step by step put together large prosopographic databases describing persons and places related to the artefacts under study. Similar trends[4] have touched most of the domains of historical research, which could hardly work without widely available well-maintained prosopographical sources.

Scholars working specifically on language have also since long developed digital databases and methodologies to deal with the huge amount of information that is required for a meaningful research in such domains as general linguistics, field linguistic or philology. Such works as that of Fr. Roberto Busa[5], producing a fully lemmatized version of the works from Thomas Aquinas, have been seminal in putting together the core methodological principles of what is now known as computational linguistics.

In this context, lexica have always played central role as major information structures for linguistic observations. Whether these are used as sources or as the main research output of a study, structured lexical data can be seen as prototypical methodological objects in the humanities. They encompass a wide range of forms, from simple word lists to complex encyclopaedic data. They intend to reach many different communities, layman or scholars. And of course, they have been, at a very early stage in the digital era, the objects of specific attention so that they could be used directly on a computer.

---

[4] See for instance the prosopography portal (http://prosopography.modhist.ox.ac.uk/) at the University of Oxford

[5] An online access to the corresponding index is available under http://www.corpusthomisticum.org/it/index.age

Observing ongoing projects working on digital dictionaries or lexical databases reflects these various aspects and the corresponding complexity in providing a coherent view upon the various type of corresponding data formats. Without aiming at being exhaustive, which would indeed be impossible given the intense activity in this domain, let us go through a few illustrative examples.

Like in many other scholarly fields of interest in the humanities, the usual transition out of the traditional book-based scholarly work is to digitize reference sources in order to increase their dissemination and usability. In this respect, many scholarly groups have focused on providing full text transcriptions of dictionaries which were either their focus of research as lexicographer, or important reference sources for wider linguistic studies. For instance, the LDI laboratory[6] has digitized several generations of the *Petit Larousse*, with the purpose of providing augmented views, field-based query facilities (see Figure 1) on all fields and cross-links with other dictionary sources (e.g. the *Dictionnaire de l'Académie*). The generic research question[7] here is to observe the evolution of structure and content across the years. As can be seen, the overall structure remains straightforward since dictionaries at this period had already acquired a good level of structural stability.

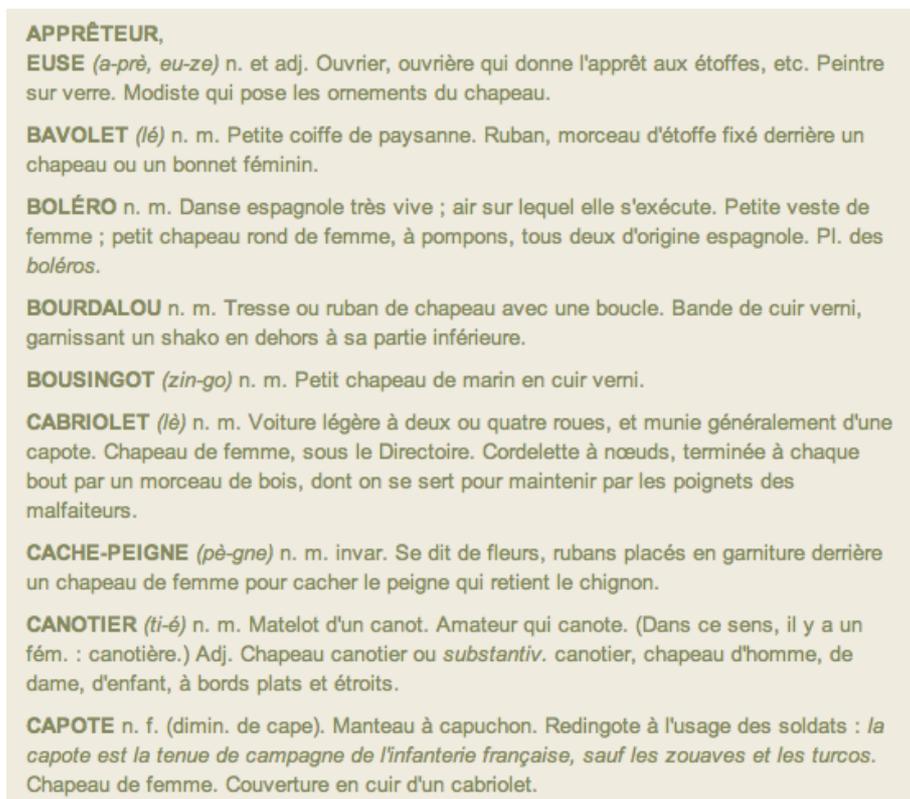

Figure 1: beginning of the results obtained for a search on "chapeau" in all definitions of the *Petit Larousse 1905*, source: LDI

On the contrary, when similar attempts are based on more ancient form of dictionaries, the achievement of a fully digitized version bumps into several kinds of

---

hurdles. Whether because of the quality of paper, the nature of the font, the orthographic variations or the actual blurred transitions between the various fields of the dictionary, it can appear to be quite complex to organise the initial material as a real structured object which can be precisely queried and cross-linked. As can be seen in Figure 2, the various levels of complexity are made even more complex because of all the implicit segments (e.g. for expressing morphological variations) that appear in even a very simple entry. As illustrated by the exemplary work done within the Textgrid project[8], the digitization of such an ancient dictionary can actually lead to the identification of generic guidelines for a wider range of dictionary type.

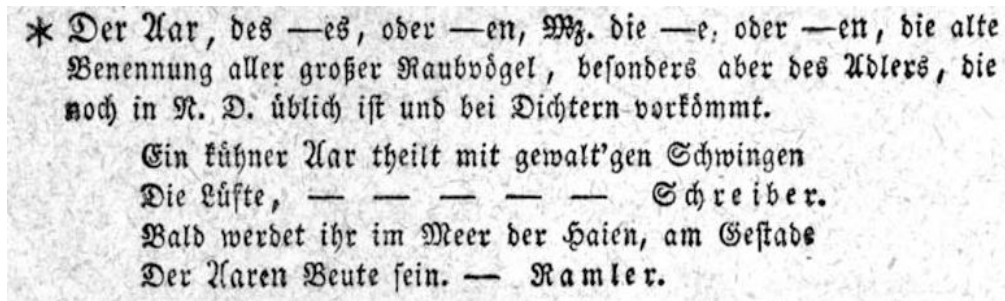

Figure 2: Entry for 'Aar' in Joachim Heinrich Campe „Wörterbuch der deutschen Sprache", 5 volumes, Braunschweig 1807 – 1811.

Beside such projects, which take a printed primary source as basis for creating the digital resource, there are more and more lexicographic initiatives that are born-digital. Some are actual dictionary projects in the traditional sense. The Franqus[9] dictionary project, for instance, aims at describing the French language as spoken in Québec and has put together a workflow and methodological principles that are very close to similar large coverage dictionary project.

Still, the recent decades have seen an increase of activity in creating lexical resources, which, instead of being intended for human usage, are conceived as information resource for the automatic processing of text and speech. Such resources do not actually contain any definition, or illustrating examples, but gather specific morphological (inflected forms) or syntactic (possible constructs associated to a word) information, which can then be used to tag the various surface forms encountered in a text. Such lexica thus bear a very flat structure, so that for a given form such as "ferme" in French, the following record can be made:

ferme ip1s ip3s sp1s sp3s im2s,fermer v

which indicates the possible morphological features associated to it (ip1s: indicative, present tense, singular, first person), the infinitive form ("fermer") and the part of speech ('v' for verb).

---

[8] Werner Wegstein, Mirjam Blümm, Dietmar Seipel, Christian Schneiker: Digitalisierung von Primärquellen für die TextGrid-Umgebung: Modellfall Campe-Wörterbuch. TextGrid-Report, 4.1, Version: 12. Oktober 2009 - 1.0 [http://www.textgrid.de/fileadmin/TextGrid/reports/TextGrid_R4_1.pdf]

[9] http://franqus.ca/projet/

Such lexica are usually joint deliverables with fully annotated corpora[10], and aim at having as wide a linguistic coverage as possible. Larger initiatives such as the Multext-East[11] consortium, have even worked in parallel on multilingual lexica covering most languages of Eastern Europe, creating the need for joint descriptive methodologies of lexical content across languages.

Finally, we would not have here a representative sample of lexical initiatives if we did not mention the importance of lexical description for the record of endangered languages and more widely for the documentation of languages and dialects around the world. In this case, the objective is no more to provide a resource which is targeted at a specific usage, but to gather as much evidence as possible about a language for which there is not in general any other recordings, in particular any written document or literature. As exemplified by the variety of projects within the DOBES initiative, the field researchers are using lexical resources as a means to combine various levels of linguistic description, ranging from the simple usage of the word in concrete utterances to the combination of multimedia illustration of the actual meaning of the word in situational contexts. As an example, Figure 1 shows how the entry for the word "tpile wee", which represents both an insect and by analogy a sung and dance performance depicting it, includes a video showing such a performance.

---

[10] A *corpus* is a collection of written or spoken documents that have been chosen or identified as representative of a linguistic behaviour and which is used to study this behaviour.

[11] http://nl.ijs.si/ME/

## Why standardizing lexical structures?

At first sight, the examples provided seem to have very little in common. Still, one should not infer that the variety of dictionary types we have observed does necessarily correspond to completely separate lexicographic practices or, even more, linguistic phenomena.

One could argue of course, that for the creator of the lexical resource, the lexicographer or linguist, a specific combination of descriptive information reflects a trade-off between the available information (e.g. field observation, extraction of lexical information from a textual corpus) and the intention to provide a comprehensive coverage of given subset of linguistic features. Depending on the actual objectives, the designer of a lexical resource may for instance focus essentially on grammatical descriptions of words, on the precise identification of senses, or the provision of testimonial examples for a corpus of literary work. By definition, such a combination of descriptive features is biased towards a certain aim, but also reflects the fact that it actually makes no sense to think of a lexical resource which would cover all aspects of linguistic description. A dictionary is the mirror of the knowledge a person, a school of thought or more widely a society has on a given language.

For this very reason, however, a given lexical resource will share with other similar, or less similar, resources some or many descriptive features, which allow one to relate and compare the available information across them. In the simplest case where the dictionary structures associate linguistic descriptions to lexical entries characterized by headwords, to describe for instance the various senses of a word, one will typically have the possibility to compare the actual orthographic forms provided for each entry, the possible grammatical features associated to each word or more specifically the domains of usage of a word. Such features, shared across several dictionaries, may allow one to see these as a continuum of lexical descriptions, which one may decide to traverse according to his or her linguistic interests.

From a technical point view, this requires that we are able to provide some level of *interoperability* between the electronic representations (the computer formats) associated with these various forms of lexical description. By providing coherent lexical formats, one may indeed intend to achieve three complementary objectives:

- To allow the export and exchange of data to third parties. In this context, interoperable data reflects the capacity to query a lexical resource and above all to parse the information provided by an external resource;
- To favour the pooling of lexical information coming from different sources, in such a way that similar pieces of knowledge can actually be mapped or at least compared;
- To limit the duplication of software development works by fostering the reusability of lexical management and consultation tools.

From a scholarly point of view, an increase of interoperability across lexical resources may also have a great impact on the capacity that scholars can have to provide fair and accurate comparisons of theirs results. There are indeed many situations where the creation of a lexical resource is an important part of the actual scholarly work of linguist. In such cases, assessing the quality of the work requires that one has the capacity to understand the organisational choices of a lexical resource and relate these to similar choices taken by others. Such a comparison may actual take place at two complementary levels. First, the one must be in the capacity to assess the general organisation of the lexical database as well as the coherence and comprehensiveness of the structure of each lexical entry. Once this is achieve, the assessment can then focus on the relevance of the elementary linguistic descriptions associated to a given entry. Without some elementary interoperability principles between lexical representations, such an assessment of lexicographic work can only be performed manually.

## But is it at all feasible?

In the most ambitious sense, it is clear that achieving full interoperability across lexical resources is just an impossible goal. This may be understood intuitively from the variety of lexical configurations that we identified so far, but also, when one takes a more precise look at it, from potential variations at various levels of representation of a lexical resource.

First, there are various possibilities in organising lexical content that may not necessarily be compatible with one another. Seen from the greatest possible distance, the main conceptual divide in lexicographic work reflects whether the word or the meaning is the actual origin of the linguistic description. A. Zauner named this difference for the first time in 1902[12] when he identified two modes of lexicographic description:

- a semasiological view of a lexicon, that associates meanings to words, as can be encountered in most usual print dictionaries. Such an organisation is usually intended for wide coverage lexica whether monolingual or multilingual.
- an onomasiological view of the lexicon, whereby words or expressions are associated to meanings, which usually correspond to a list of concept relevant for a specific field. This approach has been the core of the 20[th] century school of terminology initiated by E. Wüster[13], and is particularly suited to the description of the lexicon in specific technical fields.

Even if we remain in the semasiological trend of dictionary making, lexical entries themselves can vary a great deal in the way they are actually structured. Just to take a simple example, a lexicographer may consider a variation in lexical category (e.g. the word "cut" in English seen as a noun or verb) as strong enough a marker to justify the creation of separated entries, whereas others may just use such an information to indicate more precise sense or usage variation within an entry. The same elementary piece of information (what we will refer to later as a *data category*) can thus be considered as part of the organisational principles of a dictionary or simply as a qualifier for a specific descriptive level.

---

[12] Zauner Adolf, 1902, "Die romanischen Namen der Körperteile: Eine onomasiologische Studie", *Zeitschrift für romanische Philologie (ZrP)*. Volume 27, Issue s27, pp 40–182.

[13] Wüster, Eugen. 1991. *Einführung in die allgemeine Terminologielehre und terminologische Lexikographie*, Würzburg: Ergon.

Going deeper in the organisation of a lexical entry, we can also observe the variety of possible content values associated to specific descriptors, which can vary extensively in precision and complexity. Grammatical constructs associated to a verb can thus range from very simple opposition (transitive vs. intransitive verbs) in basic dictionaries down to complex representations of syntactic structures motivated for instance by teaching purposes[14] or adherence to a linguistic theory[15]. This may also be true for very simple features such as pronunciation where the lexicographer may want to associate precise rhythmic and prosodic information as opposed to a basic phonetic representation.

Finally, one major issue that can provide further food for thought on the complexity of lexical standardization is again the scholarly perspective. By definition, scholarly work cannot just be based on an existing fixed set of structures and categories. It should be possible at any time for a linguist to contemplate the introduction of a new category whenever he has the feeling that he has observed a linguistic phenomenon that does not match the doxa of lexicographic principles.

## Basic principles that should lead standardization in the lexical domain

In the preceding sections, we have seen the reasons for standardising lexical structures, as well as the underlying principles that should lie at the basis of such a standardisation activity. Taking a little distance, we can also identify the conditions under which we would want to see standards to emerge within the humanities community.

First, from a pure scientific and technical point of view, we need to acknowledge that standardisation is a process that should fulfil two opposite and at times contradictory purposes:

- it should be a compendium of stabilized knowledge, which documents existing practices, so that future users recognise their existing culture. IN this respect, it should cover the various variations in lexical structures that we have tried to identify in this paper;
- it should also be looking ahead for potential roadblocks so that future, and still unforeseen, forms of lexical structures could be taken in to account by means of the appropriate generalization.

Finding the best compromise between this two constraints, as well as putting together the best experts that would be able to both have an in-depth knowledge of existing practices as well as a vision of where the future of digital lexica could stay, are probably the greatest challenges that we could face in our endeavour to design optimal standards.

From an organisational point of view, we must also consider the basic constraints that bear on all standards in all fields. In this domain, identifying "real" standards boil down to the following three aspects:

---

[14] Collins COBUILD English Language Dictionary, under the editorship of John Sinclair at the University of Birmingham.

[15] Cf. the adoption of Mel'cuk linguistic theories in: Binon Jean, Serge Verlinde, Jan Van Dyck, & Ann Bertels, *Dictionnaire d'apprentissage du français des affaires*, Editions Didier, Paris 2000, (ISBN 2-278-04356-0)

- as anticipated, a standard should be the result of a *consensus* building process. As a result, a standard proposal can by no means be self claimed by an individual or a group without a prior check that it bears acceptance from a wider community. In general, in particular in ripe organisation such as ISO or the TEI, standards are iteratively design in order to fulfil this criterion;
- a standard should also be *available* in the long term, so that a stable reference to it can always be made and consequently provide a stable background for many possible application;
- finally, a standard should be *maintained*, so that technology changes or progress within a community be systematically incorporated in new versions of the document. This maintenance is usually based upon either a systematic review (as carried out within ISO) or a feedback mechanism reporting bugs or requiring new features to the standardisation body (as in the TEI).

These three constraints make a core characterisation of the kind of services a standardisation body should offer and *a contrario* what proprietary initiatives should be aiming at when wanting to dissemination a given specification as an international standard.

### Can scientists bear standards?

Even though we have managed to give quite an extensive account on why standards are essential for scholarly work in (at least) the linguistic domain, tackling at times the actual difficulties that such an endeavour may be facing, it can also be interesting to identify why scholars may just be reluctant in adopting standards as a core component of their lexicographic work. In the following section, which has to be read with the appropriate distance, we try to identify good or lesser good reasons in this respect.

### Freezing knowledge

Standards are basically associated with the notion of stability. By definition, a standard is a fixed set of constraints that one has to follow in order to be compliant with it. As such, it is understandable that standardizing could be seen as an activity that is contradictory to the notion of scientific discovery, especially if the topics at hand are closely related to domains where scholarly work is very active. In such cases, we need to find ways to either show the complementarities between what is being standardized and what is the object of research, or even, provide means to continuously integrate new concepts in the process of updating standards.

### Losing one's time

Complying to a standard, not to mention contributing to standards development, can be seen as a loss of a precious time that would be otherwise dedicated to research proper. When, in the course of a specific research project, one has to produce some digital primary sources, or an electronic version of a lexical resource, it is often seen as much more efficient to just improvise a self-defined computer format then spend time in reading a documentation that may all in all not even match the precise needs of the research. After all, scholars know exactly what they need, don't they?

### Forcing diverging views to agree

One of the most complex issues in standard development is to find an optimal consensus between technical views that may initially differ a great deal. This is all the more sensible in the humanities where in particular the names and precise definition given to concepts is strongly related to specific schools of though. For instance, there is hardly any agreement on how to name the arguments of a verb, since it is strongly related to the understanding that a scholar may have of the corresponding semantic relations[16]. This aspect of standardisation is all the more problematic when a specific scholar or school of thought feels it has not be involved in the standardisation process and may reject in principle the resulting proposal. Anyone coordinating a standardization activity within the humanities should definitely keep this in mind and make sure that a) the group of experts involved in the definition of the standard is representative of the corresponding community and b) there exist mechanisms to account for various school of thoughts within the standard.

### Forcing one to make data accessible to others

Even if the argument is not explicitly made, experience proves that non-compliance to standards is often seen as a way to ensure that scholarly data can hardly be used without involving the initial producer. There is always a fear that by giving up to others may lead some to publish results even quicker than those who have issued these data. On the contrary, standardisation can be seen as a facilitating factor to get data samples easily from a given producer to another party interested in it. In the humanities, as in any other kind of science, this is an essential factor to progress.

### A crash-course in XML

Before we go any further with this paper, it is necessary to provide some background technical knowledge to our reader. Indeed, we have been speaking a lot in the preceding sections about lexical structures and formats without actually hinting at the way such structures could concretely be represented in a computer. Besides, it just happens that in the current web-based technological context a representation language has taken the lead for all of data interchange on the Internet, namely XML, the eXtended Markup Language, and this language has been the basis for the definition of most standards that are applicable in the linguistic domain. As a result, we allow ourselves to convey here the minimal knowledge about XML, so that the reader may fell, if not at ease, minimally acquainted with the vocabulary we will be using later on in the paper.

### XML is about trees

XLM was designed to account for the presentation of tree structures. From a computational point of view such a tree structure (cf. a simple example in figure XX) is characterized by one single *root* node (here *gramGrp*) and each node other than the root node has one and only one *parent node*.

---

[16] Cf. ISO WD 24617-4, Language resource management — Semantic annotation framework — Part 5: Semantic Roles (SemAF-SRL), edited by Martha Palmer, University of Colorado Boulder.

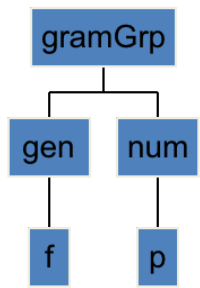

Figure 4: a simple tree structure (grammatical information in TEI)

## XML is about angle brackets

The very simple structure delineated above can be represented (or *serialized*) in XML as a sequence of characters organized as *elements,* which can be isomorphically (i.e. one-to-one) mapped onto nodes of a tree. An element is identified by means of an *opening tag* (e.g. *<gen>*) and a *closing tag* (e.g. *</gen>*). To reflect on the hierarchical structure of a tree, element may be embedded one within the other (the element *gen* is embedded in the element *gramGrp*) but can in no way overlap. The following XML excerpt serializes the tree presented in the preceding section:

```
<gramGrp>
        <gen>f</gen>
        <num>p</num>
</gramGrp>
```

## Issues with using XML for data modelling

In the XML context, modelling data structures in general resides in carrying out two complementary activities:

- Defining the vocabulary of tags that are allowed to describe such a structure, as well as the rules for combining them, i.e. which element is allowed within which context and how many times;
- Providing the actual semantics associated with each element, in order to ascertain to which purpose a given element can be used and under which conditions.

For instance, in the above example, one should be able to express that the <gen> element should occur within the <gramGrp> at most once and that indeed, <gen> indicates the grammatical gender within a group (<gramGrp>) of grammatical features attached to a lexical entry.

It is essential that such syntactic and semantic constraints, even if instantiated within the specific technical framework of the XML language, be accessible to scholars, since at the end of the day, they should precisely reflect the conceptual structures that the scholar has identified as relevant for his field of knowledge.

## From theory to practice – standardisation in the open sea

The theory so far was nice and actually covers all aspects that anyone would have in mind in a similar context. However, all this would probably be completely useless if we did not have the prospect to actually have real standards being designed. Since there are indeed some, we will now see how such standards are actually designed in the light of there specific standardisation context, since we will address, in the coming sections, the work done by two complementary institutions:

- The Text Encoding Initiative (TEI), which has been put together within the humanities themselves and attempts to provide ready to use formats for the representation of digital texts;
- The International Organisation for Standardisation (ISO), which is an international consortium, covering all technical domains and which has recently been pursing work on the standardisation of language resources, in its committee TC 37/SC 4.

By describing how both initiatives have managed to deal with the lexical issue, we will try to both identify which kind of purpose they thus aim at and see how much, though being complementary in nature, they should both be seen as modelling tools for scholars.

### Modelling lexical structures with the TEI

The Text Encoding Initiative was initiated in 1987 when a group of textual databases worldwide decided to join efforts and define guidelines for the representation and interchange of textual documents. As the year of their first meeting immediately followed that of the publication of the ISO SGML standards (which was to become XML 15 years later), the consortium unanimously agreed to adopt it as background for the definition of their own recommendations[17].

One of the main characteristic of the TEI infrastructure is that, while being an XML application, it is not intended to be used on-the-shelf as a monolithic group of XML objects. Indeed, the TEI is by construction an environment that a user needs to adapt (customise) to his/her own needs when using the TEI for a specific type of documents.

The customizability of the TEI results from its general information architecture that is based on the following core concepts:

- the TEI guidelines is based on a series of *modules*, each representing either generic technical components (core elements, header, encoding of names) or domain specific subsets (for the encoding of specific genres or document types, e.g. drama, manuscripts or the transcription of spoken data. The module for "print dictionaries" is one of the later type;
- most of the elements are attached to classes which group together those with similar semantics or having the same structural behaviour (e.g. appearing as children of the same elements).

The organisation of the TEI ontology in modules and classes is particularly important to allow the appropriate selection of descriptive elements that may fit a particular

---

[17] The work carried out in defining the TEI guidelines actually contributed to the emergence of XML, with TEI experts such as Michael Sperberg-McQueen, Steve DeRose or Henry Thompson taking the lead in defining many of its core concepts. [Lou, Seb Alan Renear]

purpose. In the following section, we will see some of the concrete possibilities through the presentation of a simple example.

## Matching the TEI "standard" with scholarly needs

Let us consider a scholar who would want to describe a simple dictionary using the TEI infrastructure. He/she will first select the basic modules of the TEI, allowing him to have the generic structure of a TEI document comprising meta-data (the TEI header), together with the "dictionary" module, that will provide him with the various elements he/she may need for representing a lexical entry. By doing so, he/she will immediately be able to set-up an environment where he can write constructs such as the one represented in Figure 5. What the TEI provides him/her with at this stage is the following:

- a group of elements to represent lexical features within a dictionary entry, for instance <pos> for representing the grammatical category ("part of speech") of a word;
- the syntactic constraints bearing upon these XML elements, by means of an XML schema (DTD, RelaxNG or W3C schema);
- the corresponding semantics, expressed by means of a precise online documentation, combining a comprehensive description of the way dictionaries may be represented[18], and a specific documentation for each element[19].

```
<entry>
 <form>
     <orth>table</orth>
 </form>
 <gramGrp>
   <pos>n.</pos>
     <gen>f.</gen>
 <gramGrp>
 <def>Pièce de mobilier…</def>
 <cit>
     <quote>Une table de cuisine</quote>
   </cit>
</entry>
```

**Figure 5: A simple (constructed) dictionary entry represented according to the TEI guidelines**

Once such a first representation is achieve our scholar will then want to refine his/her representation in various ways and make sure that these constraints are reflected in the schemas and documentation he/she will give to his/her students to produce further lexical entries. Let us see how the TEI mechanisms would allow him/her to actually implement three types of constraints.

---

[18] http://www.tei-c.org/release/doc/tei-p5-doc/en/html/DI.html
[19] for instance, the documentation for the <pos> element under: http://www.tei-c.org/release/doc/tei-p5-doc/en/html/ref-pos.html

### Selecting appropriate descriptors

As seen in the various examples that we presented earlier in this paper, the actual combination of possible features for the various components of a lexical entry is an essential design aspect of a lexical database, in particular when such features are not at all relevant for a given language. Typically, a human oriented dictionary will limit grammatical information to a set of very basic features related to the provision of the grammatical category (<pos> – *part of speech* element in the TEI) and possibly grammatical gender (<gen> – *gender* element) for nouns and transitivity (<subc> - *subcategorization*) for verbs. Through its class system, such a de-selection is a core mechanism of the TEI infrastructure, since each element is individually connected to a class. For instance, all grammatical descriptors are members of a single class (named "model.gramPart") and a schema tuned for a specific project could actually limit the content of this class to <pos>, <gen> and <suc>, whereas other elements available in the TEI vocabulary (e.g. <mood>, for expressing the grammatical mood of verbs) would be disallowed.

### Adding one's own categories

In complement to this first customisation possibility, the TEI allows one to define its own extension to the existing elements. This is particularly needed in a generic framework such as the TEI to account for descriptive features that a scholar may want to express, but which are not consensual enough – or specific to a group of languages that have not been represented in the standardisation process - to be integrated into the standard. For instance, a lexicographer that would want to represent a language that bears politeness markers on inflected of verbs (e.g. Japanese or Korean), could document an additional element named <politeness> and make it a member of the class of elements forming part of a grammatical description (<gramGrp>).

### Constraining possible values

In many projects, it is important to set constraints on the possible values of a descriptive element, so that, for instance, the various editors involved in a dictionary projects do not provide a grammatical gender as a random set of strings such as: "f", "fem", "feminine", etc. To this end, the TEI infrastructure offers a "change" mode in the specification of a customized schema, so that an element such as <gen> (*gender*) keeps all its characteristics, except for what the user has explicitly modified, for instance setting the possible values to the set {m,f}.

### Overview

The TEI infrastructure is set in such a way that by default a user has a usable infrastructure and can, with little or no extra effort, use it directly to encode its digital data. Still, the specificity of a dictionary project can be taken into account by customizing the infrastructure and in this respect, the TEI offers simple mechanism not only to express constraints on the available XML objects, but also to benefit from a comprehensive documentation for this new schema. This in turn contributes to a better exchange of information about practices between scholars.

For some scholars, however, it can be seen as an overhead not to have a more abstract modelling tool at hand that would not be necessarily tied to a fixed XML vocabulary. We shall see in the following sections in which way ISO provides such environment.

## Lexical descriptions in the ISO context

The International Organization for Standardization (ISO[20]) is one of the major standardization bodies worldwide. By and large, its coverage encompasses most technical and scientific domains and it can be seen from two complementary perspectives:

- from an administrative point of view, ISO is an association of national standardization bodies which each have an equal right to initiate, comment and approve a standard project;
- when looking at the technical content proper, ISO is organized in technical committees and sub-committees which group together experts within a dedicated field where the standardization work is to take place.

Following a whole series of international funded projects[21] in the 90s aiming at providing (pre-normative) guidelines for the encoding of language resources, various experts worldwide deemed it necessary to go a step beyond and initiated the creation of a new ISO committee (ISO/TC 37/SC 4) dedicated to language management issues. This committee, grounded in 2002, quickly gained success and put together a whole portfolio of standards and standard proposals covering most domains needed for achieving interoperable language resources[22].

## The ISO "Lego" model of lexical structures

ISO committee TC 37/SC 4 started to work in 2003 on offering a standardisation of lexical structure in the context of the LMF (Lexical Markup Framework) project. Published in 2008, ISO standard 24613 exemplifies perfectly the modelling strategy developed for language resource modelling at large[23][24].

The modelling framework adopted by ISO committee TC 37/SC 4 is inspired by information modelling principles developed within the object orient languages and implemented by the OMG in the UML specification language. The framework requires one to describe an informational structure on the basis of two complementary elements:

- a *meta-model* which represents the abstract organisation of information and informs current practice for a given type of information. Such a meta-model is described as a combination of *components*, representing elementary units of information ;

---

[20] http://www.iso.org

[21] In particular EAGLES (http://www.ilc.cnr.it/EAGLES/home.html), initiated by Prof. Antonio Zampolli

[22] See http://www.iso.org/iso/iso_catalogue/catalogue_tc/catalogue_tc_browse.htm?commid=297592&published=on&development=on for an overview of the program

[23] Romary L., Salmon-Alt S., Francopoulo G., Standards going concrete: from LMF to Morphalou", 20th International Conference on Computational Linguistics - COLING 2004 (2004) - http://hal.inria.fr/inria-00121489/fr/

[24] Romary L., "Standardization of the formal representation of lexical information for NLP", *Dictionaries. An International Encyclopedia of Lexicography. Supplementary volume: Recent developments with special focus on computational lexicography*, Mouton de Gruyter (Ed.) (2010) - - http://hal.archives-ouvertes.fr/hal-00436328/fr/

- a selection of *data categories* that can be used to qualify each component of the meta-model, and which form the basis for instantiating a meta-model.

In its simplest form, the LMF metamodel can be depicted as in Figure 1, where a lexical database is seen as a combination of some metadata (*Global Information*) and a series of lexical entries. According to the semasiological view, each lexical entry is characterized by means of a form, which in turn may be associated with one or several meanings. As can be seen, meanings can be recursively embedded to form a full semantic description of a word. The diagram also shows that a lexical model can be further designed by adding lexical extensions to the core metamodel so that specific representations dedicated to syntactic or multilingual descriptions can be added as needed to shape a more complex lexical metamodel. In order to have a full lexical model, such a metamodel must be combined with a selection of descriptors, also known as data categories.

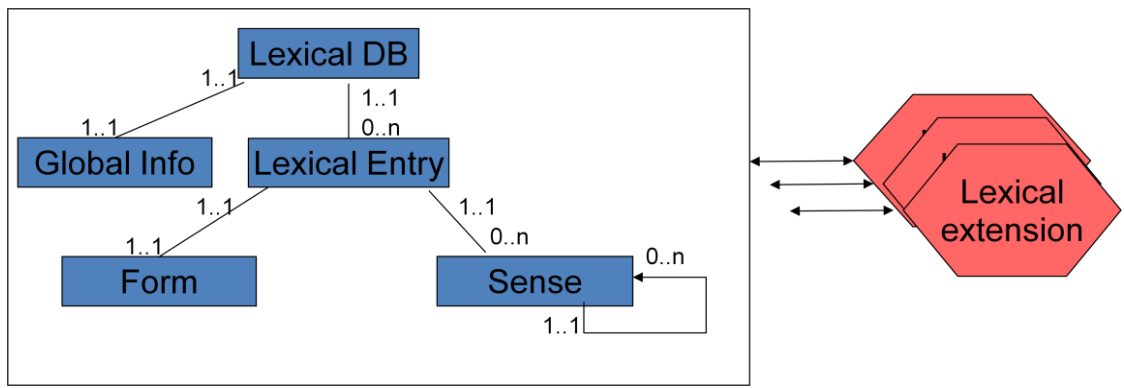

**Figure 6: The LMF (ISO ) metamodel.**

To illustrate this modelling process, we can outline a possible model of full form lexica that may be used to describe the inflected forms of a given language. To this purpose, we consider that each lexical entry groups together all its possible inflected variations within a specific extension whose entry point is the component *Morphology*. This component contains in turn an optional *Paradigm* component to characterize the inflection class of the lexical entry (for instance, first group of French verbs with "–er" ending) and as an *Inflexion* component, which can be iterated as many times as needed. The combination of these components with the core LMF components leads to a structure depicted in Figure 7.

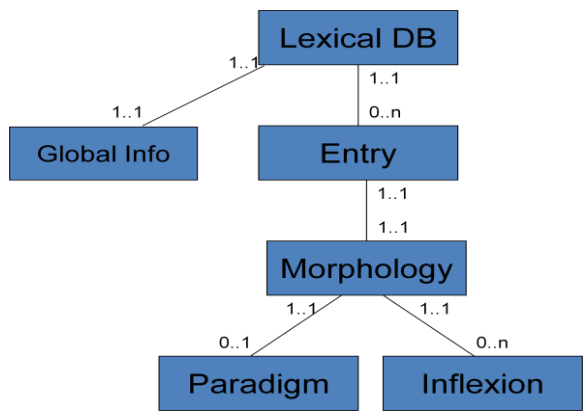

**Figure 7: An LMF meta-model for full-form lexica**

This general structure, extending the core LMF meta-model, is in turn a meta-model for any language for which such a description could be applied. Still, if we actually want to make it a complete model for a full-form lexicon, we now need to "decorate" it by means of data categories corresponding to the characteristic of the language to be represented. This is illustrated in Figure 8, where a simple model has been outlined based on three groups of data categories, namely:

- the lexical entry is characterised by a lemma and a grammatical category (*part of speech*);
- the paradigm is simply associated to a paradigm identifier;
- each inflected form is described by means of an actual written form (*word form*) and a series of grammatical characteristics for gender, number, person or tense.

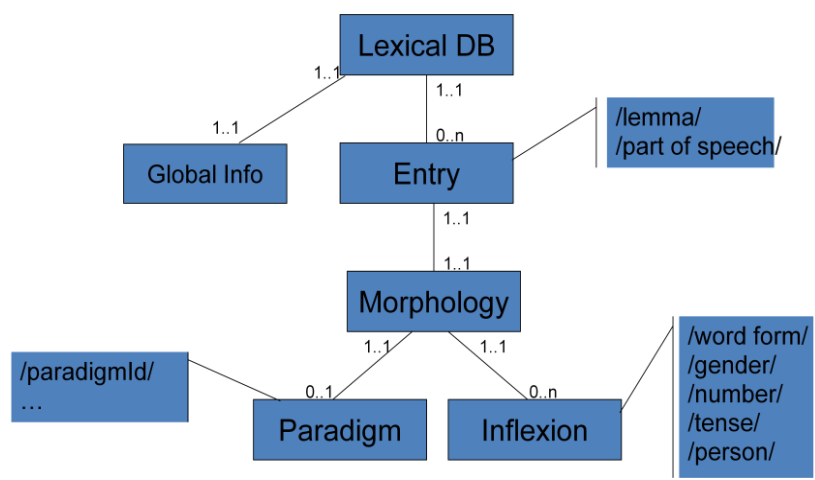

**Figure 8: A completed LMF compliant model for full-form lexica.**

Even if this model is particularly simple, it is already sufficient to describe full-form lexica in quite a number of western languages. According to the LMF principle such representations can be expressed in XML using any kind of vocabulary, under the condition that the XML abstract structure be isomorphic to the model outlined in Figure 8. For instance, Figure 9 shows the entry that would correspond to the word "chat" (*cat*) in French.

```
<lexicalEntry>
    <lemma>chat</lemma>
    <grammaticalCategory>noun</grammaticalCategory>
    <morphology>
      <paradigm>
          <paradigmIdentifier>fr-s-plural</paradigmIdentifier>
      </paradigm>
      <inflexion>
          <wordForm>chat</wordForm>
          <number>singular</number>
      </inflexion>
      <inflexion>
          <wordForm>chats</wordForm>
          <number>plural</number>
      </inflexion>
      …
    </morphology>
</lexicalEntry>
```

Figure 9: Sample entry for a full-form lexicon.

## Overview

As can be seen, ISO standard 24613 (LMF) provides a very powerful tool for anyone to design all kinds of lexical structures as needed in scholarly or commercial contexts. Its flexibility could indeed also be seen as a handicap since it allows one to create combinations of lexical information that may not necessarily make sense from a lexicographic point of view. Still, in the same way as the TEI does, the elicitation of core concepts and the corresponding methodology creates a joint culture across lexical projects so that they can further work together on joint (LMF compliant) guidelines.

Another issue to consider here is of course that LMF does not provide a fixed XML serialisation, which prevents it from being used as a real interoperability standard. Each application or group of applications must identify which serialisation optimally represents the corresponding model. Clearly, one could think, in the perspective of further convergence across standardisation initiatives, to assess how much the TEI and its customization facilities could play a role to this end.

Finally, we have so far been using the concept of data category without providing a precise account of their nature and role in the standardisation process. This will be the theme of the last part of this chapter.

## Data categories as a conceptual market place

Data categories are elementary descriptors seen as abstractions upon the various possible implementation as a database field, an XML object, or whatever human- or machine-readable representations. For instance, a field linguist may define a series of simplified codes while transcribing and annotating some recordings on paper and in parallel associate such codes with reference data categories. This will allow him/her to document his/her data and make sure he/she can compare them with the work of others, or even be able to work again on his own observations in the future. The same applies if he/she has recorded part of his data in a computer file, whether word processor, spreadsheet or more elaborated database.

In order to fulfil the needs of describing basic feature-value representations, data categories can be of two basic sorts:

- *complex data categories,* corresponding to placeholders for a specific descriptors, such as /part of speech/, /grammatical gender/ or /author/;
- *simple data categories*, which represent elementary values such as /feminine/, /plural/, or /ablative/.

Complex data categories can naturally be constrained by providing either a generic data type (e.g. date, number, string, etc.) or a list of allowed simple data categories.

As such data categories plays two complementary and closely related roles:

- they are a tool to record and document the semantics of the concepts used by a scholar in the course of his data description activities;
- they provide means for an accurate specification of data formats, which are in keeping with the actual scientific concepts that the scholar has mind with regards his data.

This last point is particularly important to bear in mind since it represent a ground basis for the actual interface between a scholar and the information technologist that may be in charge of providing him with the appropriate tools to fulfil his data based research. A remarkable example of such a configuration is indeed the work done at the Max Planck Institute for Psycholinguistic in Nijmegen, where the Lexus tool allows the linguist to specify (in an LMF compliant way) his lexical structure, before he actually starts entering his data in this lexical management environment.

To formalise further the notion of data category, we can refer to the standardisation work carried out within the database community to record the semantics of fields in a database model[25]. As depicted in Figure 10, ISO 11179 formalises the notions that we outlined at the beginning of this section as a two-level model:

- The first and abstract level organises information objects at *data element concepts*, which in turn may me characterised by means of a conceptual domain;
- The second level corresponds to the concrete instantiation of the first level within an information structure and is based upon *data elements*, which in turn may have values taken out of a *value domain*.

The one difference between the ISO 11179 model and the notion of data categories described here is that *simple data categories* do not have real equivalents in ISO 11179. This is indeed a major addition that was to be made since elementary values such as /singular/ are core members of a linguistic ontology.

---

[25] In ISO/IEC joint committee JTC1/SC 32

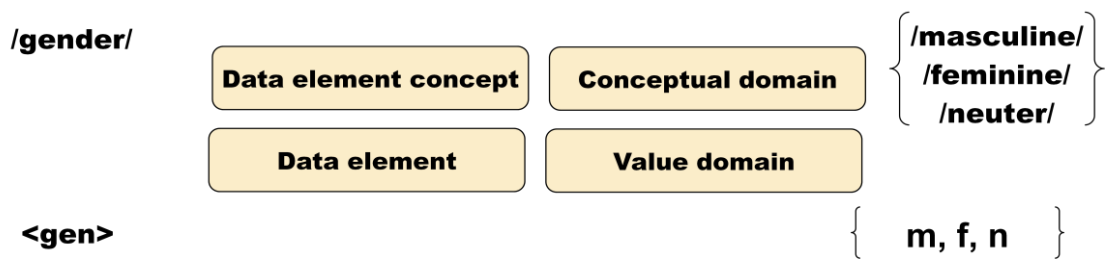

**Figure 10: The ISO 11179 organisation.**

## The limits of standardisation – the case of gender

Providing such an abstract model as the one presented before may not suffice to satisfy representational needs of scholars working on languages. Indeed, it is probably the right time for us to stay away for a while from technological aspects and ponder upon the possibility to standardise at all a given linguistic concept. As can be anticipated it is already strong debate, which we can just tackle here at the surface by providing some general lines of tensions as well as possible grounds for compromise in the context of a specific example.

The data category we will consider here is that of *grammatical gender*, which is probably one of the most widely used descriptor in lexicographic works. It is for instance a core component of any *tagset*, that is the list of descriptive tags used by computational linguists to annotate a text at the word level. As such many implementers of such natural language processing systems have consider since many years[26] that providing a reference description for such a notion would be essential.

At the opposite, one could assert – the view that prevails for typologists – that no two single concepts are shared between two given languages, even within one language between two observations. The idea is that a descriptive feature may always be seen as the specific link between a language sample and the analysis of a phenomenon, thus forbidding any kind of generalisation.

To better situate where the problem stays, let us consider the issue of /gender/ as a possible candidate for standardisation. The category has many interesting feature as an illustrative example. It is intuitive enough for most languages, is widely used in many lexicographic applications and as we shall see has the appropriate complexity to understand where standardisation can find its place.

As defined in by G. Corbett[27], grammatical gender is a purely morphosyntactic concept seen as "a classification of nominals, as shown by agreement". Such an agreement is usually related to the association of determiners with nouns as can be observed in the disctinction in German between "die Katze" (*the cat*; feminine) and "der Hund" (*the dog*; masculine). Such an agreement can then propagate within a sentence to various linguistic components such as adjectives, numerals or verbs.

One thing at least is clear at this stage, the notion of gender does not match at all that of natural sex, and any further analysis should be strictly based on linguistic grounds and observation. This point definitely rules out, when contemplating standardisation, any further use of generic database oriented standards such as ISO 5218[28], which is intended to represent human sex in computer applications.

---

[26] See for instance the work done in the context of the Multext-East project: http://nl.ijs.si/ME/
[27] Greville G. Corbett, "Number of genders", wals.info

The definition of gender in relation to agreement can also be expressed over wider distances of texts when separate pronouns[29] may be used anaphoricaly in relation to a preceding noun phrase of a given gender. This may lead to elaborate pronoun systems as illustrated in Table 1 for Rif Berber, where gender distinction apply to both second and third person pronoun both with singular and plural genders.

| 1sg | nəš | 1pl | nəšnin |
|------|------|------|--------|
| 2sg.m | šək | 2pl.m | kəniw |
| 2sg.f | šəm | 2pl.f | kənint |
| 3sg.m | nətta | 3pl.m | nitnin |
| 3sg.f | nəttæθ | 3pl.f | nitənti |

**Table 1: personal pronoun system in Rif Berber (McClelland 2000: 27)**

Another issue is that both the notion of gender itself, as well as the number of possible values for gender varies dramatically from a language to another. Whereas gender agreement does not exist at all in some languages like Japanese (and with a very limited scope in English), some languages appear to have a complex gender system with sometimes more than 20 values[30]. Such an observation in itself could just jeopardize any attempt to standardize the notion of grammatical gender at all, but before yielding out, we may want to consider the issue further.

Indeed, we would like to make two complementary considerations as to what "standardising gender" would mean. First, whether or not it applies to a specific language, gender as a grammatical concept, is widely shared across scholars or engineers working on languages at large, whether they describe particular languages or implement tools for analysing them. As such, it may be important to provide them with a fixed point to which they could systematically refer, when they want to make sure that colleagues, or other software systems, will understand their data in the same way. This would for instance facilitate cross-language studies, the definition of generic query systems for linguistic corpora, or the design of similar presentation modes for online language learning environments. From this point of view, "grammatical gender" as such, but also probably elementary values for gender, could be given standardised identifier, together with some generic, if possible language-independent, definitions. Moreover, the variations across languages as to how and under which conditions grammatical gender applies, could and probably should be recorded in order to a) refine the generic definitions provide for the concept at large and b) to provide further constraint, through the precise listing of applicable values. Of course, this second level leads to a potentially complex attempt at precisely describing the languages of the world and one must provide the means to see this as a long-term endeavour.

---

[28] –ISO 5218, Information technology — Codes for the representation of human sexes. This standard indentifies four values, namely : 0 for 'not known', 1 for 'male'; 2 for 'female' and 9 for 'not applicable'

[29] Anna Siewierska , "Gender Disctinctions in Independent Personal Pronouns", wals.info (http://wals.info/feature/44)

[30] Greville G. Corbett, "Number of genders", *wals.info*.

To summarise, the example of grammatical gender can be seen as a prototypical case of the complexity of standardising linguistic concepts in general. Still, we do think that the endeavour is manageable if we offer a standardisation framework implementing a good compromise between linguistic genericity and linguistic accuracy. Such a framework must allow both engineers and scholars to feel at ease by finding there both stable points of reference and trustful linguistic content.

## The ISO data category model

In the context of its work on modelling various types of language resources, ISO technical committee 37 has designed a specific framework for recording, documenting and standardizing data categories. Among the design principles of the *data category registry*[31], we can mention here the following guiding ideas:

- providing an open space of reference concepts allowing linguists and developers to relate his own practice with standardized definitions and identifiers;
- providing an extensive multilingual support so that both the variation of the semantic of a category across languages and the recording of terms used to refer to them would be precisely taken into account;
- complying with existing standards and practices such as ISO 11179 for metadata registries (used in the field of computer databases) or the W3C OWL recommendation (for representing ontologies).

In order to provide the best compromise to these constraints, a two level model was introduced, as depicted through the specific example of /grammatical gender/ in Figure 11. The first level of representation, which is also the entry point for the category, provides a unique identifier for persistent reference, a generic definition (in one or several languages) for this category, one or several profiles (the possible domains of application), and, when applicable, a list of possible values for the data categories. These values, recorded in the so-called *conceptual domain*[32] of the data category, are the set of all recorded values, independently of language-specific constraints.

[31] Ide Nancy & Laurent Romary. A Registry of Standard Data Categories for Linguistic Annotation. 4th *International Conference on Language Resources and Evaluation - LREC'04*, 2004, Lisbonne, Portugal, pp. 135-138, 2004.
[32] As defined in ISO standard 11179 part 3.

<table>
<tr><td><em>Entry Identifier</em>:</td><td style="color:red">grammatical gender</td></tr>
<tr><td><em>Profile</em>:</td><td>morpho-syntax</td></tr>
<tr><td><em>Definition</em> (fr):</td><td>Catégorie grammaticale reposant, selon les langues et les</td></tr>
<tr><td></td><td>systèmes, sur la distinction naturelle entre les sexes ou sur</td></tr>
<tr><td></td><td>des critères formels (Source: TLFi)</td></tr>
<tr><td><em>Definition</em> (en):</td><td>Grammatical category … (Source: TLFi (Trad.))</td></tr>
</table>

| *Object Language*: fr | *Object Language*: en | *Object Language*: de |
|---|---|---|
| *Name*: genre | *Name*: gender | *Name*: Geschlecht |
| *Conceptual Domain*: {/feminine/, /masculine/} | *Name*: grammatical gender | *Name*: Genus |
| | | *Conceptual Domain*: {/feminine/, /masculine/, /neuter/} |

**Figure 11: The grammaticalGender data category (constructed example)**

The second level of representation is dedicated to language specific information and may be iterated as many times as there is a need for it. Two essential types of information may be recorded there: first, it is the place where any refinement of the semantics of the data category for the corresponding language can be traced, for instance by means of a more precise definition, or as is the case in the example, by indicating a subset of the possible values applicable for that language (grammatical gender in French can only take two values, namely, masculine and feminine). Second, the actual terms that may be used in the language at hand to refer to the data category are listed, to facilitate search and/or display of the entry.

As can be seen, the model designed by ISO (and implement within the ISOCat platform[33]) is inherently semasiological, since it places the data category, as concept, at the centre of information recorded by the registry. Even more, it goes beyond the traditional notion of semasiology, which is limited to lexical description, to encompass the provision of multilingual semantic constraints for a concept.

## Standardisation as a component of the scholarly process

As could be induced from the sole evangelising tone of this paper, we are only at the start of a process where standardisation becomes a natural dimension for scholarly work in the humanities. Whereas specific sub-communities have since long integrated this component, typically for annotated text representation within the Text Encoding Initiative, it remains a non-consensual issue for many scholars whose perspective is not immediately to be able to interchange data with colleagues. Still, as depicted in Figure 12, we want to conclude by arguing that the standardisation process belongs to the knowledge discovery endeavour. Thereby, at some specific stages, a scientific community identifies and references what in its everyday practices may be seen as stable knowledge that could be widely disseminated as standardized concepts. Such standardized concepts, as they are extensively used and implemented outside the initiating scholarly context, may be put to the test as well as used within the scholarly world itself as comparison points for further analyses or experiments.

---

[33] http://www.isocat.org

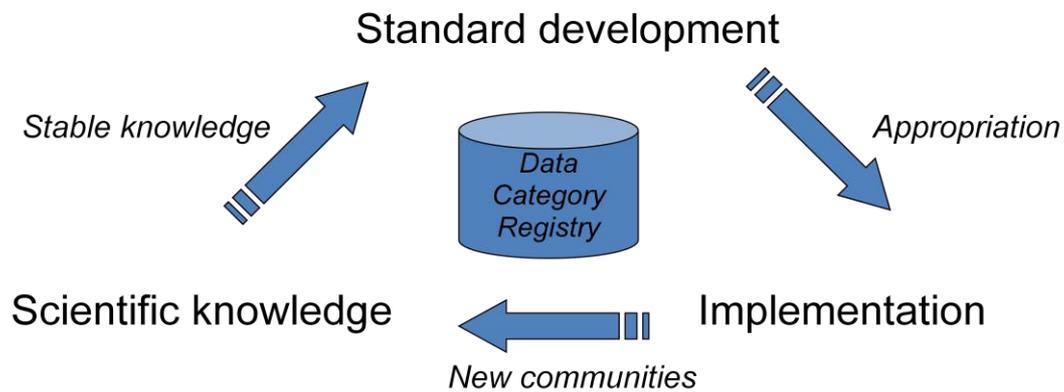

**Figure 12: Relating standards creation with the scholarly process.**

# Acknowledgements

The author wishes to address his warmest thanks to the various colleagues who have provided him with additional illustrative examples or have taken the time to comment on this paper. Specific thoughts are expressed to Werner Wegstein, Hélène Manuelian and Marc Kemps-Snijders.